\documentclass[sigconf]{acmart}
\AtBeginDocument{%
  }

\setcopyright{acmlicensed}
\acmISBN{978-1-4503-XXXX-X/2018/06}

\usepackage{multirow}
\usepackage{booktabs}
\usepackage{float}
\usepackage{xcolor}
\usepackage{mathtools}
\usepackage{amsmath} 
\usepackage{bm}
\usepackage{geometry} 
\usepackage{booktabs} 
\usepackage{tabularx} 
\usepackage{caption}  

\usepackage{graphicx}
\usepackage{xcolor}
\definecolor{LightGreen}{RGB}{46,125,50}    
\definecolor{LightOrange}{RGB}{245,124,0} 

\usepackage{CJKutf8} 

\usepackage[unicode]{hyperref} 
\usepackage{bookmark}

\begin{document}
\begin{CJK*}{UTF8}{gbsn} 
\title{LiveThinking: Enabling Real-Time Efficient Reasoning for AI-Powered Livestreaming via Reinforcement Learning}

\author{Yuhan Sun}
\email{yuhansun.syh@taobao.com}
\affiliation{%
  \institution{Taobao \& Tmall Group of Alibaba}
  \city{Hangzhou}
  \state{Zhejiang}
  \country{China}
}

\author{Zhiwei Huang}
\email{xuanan.hzw@alibaba-inc.com}
\affiliation{%
  \institution{Taobao \& Tmall Group of Alibaba}
  \city{Hangzhou}
  \state{Zhejiang}
  \country{China}}

\author{Wanqing Cui}
\email{cuiwanqing.cwq@alibaba-inc.com}
\affiliation{%
  \institution{Taobao \& Tmall Group of Alibaba}
  \city{Beijing}
  \country{China}}

\author{Shaopan Xiong}
\email{xiongshaopan.xsp@alibaba-inc.com}
\affiliation{%
  \institution{ROLL Team of Alibaba}
  \city{Hangzhou}
  \state{Zhejiang}
  \country{China}}

\author{Yazhi Guo}
\email{guoyazhi.gyz@taobao.com}
\affiliation{%
  \institution{Taobao \& Tmall Group of Alibaba}
  \city{Hangzhou}
  \state{Zhejiang}
  \country{China}
}

\author{Meiguang Jin}
\email{meiguang.jmg@taobao.com}
\affiliation{%
 \institution{Taobao \& Tmall Group of Alibaba}
 \city{Beijing}
 \country{China}
}
\authornote{*Corresponding Author}

\author{Junfeng Ma}
\email{jack.majf@taobao.com}
\affiliation{%
 \institution{Taobao \& Tmall Group of Alibaba}
 \city{Hangzhou}
 \state{Zhejiang}
 \country{China}
}


\begin{abstract}

The real-time nature of e-commerce livestreaming requires digital avatars to respond immediately to viewer comments, ensuring conversational fluency and seamless user experience. This goes beyond typical chatbot functionality, requiring not only correctness and helpfulness but also ultra-low latency. This poses a significant challenge for state-of-the-art Retrieval-Augmented Generation (RAG) systems powered by Large Reasoning Models (LRMs), which often incur high inference delays despite their strong reasoning capabilities.
To mitigate this latency-quality trade-off, we propose LiveThinking, a two-stage optimization framework. First, we employ knowledge distillation via Rejection Sampling Fine-Tuning (RFT) to compress a large teacher LRM into a lightweight architecture, significantly reducing computational cost. While this drastically lowers inference cost, the distilled model inherits verbose reasoning paths that still exceed latency budgets. In the second stage, we apply reinforcement learning via Group Relative Policy Optimization (GRPO) to directly shorten the reasoning path, with a reward function that prioritizes response brevity while preserving correctness and helpfulness.
LiveThinking demonstrates strong performance on both public benchmarks and real-world deployments. On the multi-hop reasoning dataset MuSiQue, our 30B MoE model (3B active) outperforms a 670B teacher in EM (+12.3) and F1 (+10.2), with 41\% shorter responses and 95\% lower decoding cost. Deployed on Taobao Live, LiveThinking reduces computational cost by 30× (12× from distillation and 2.5× from reasoning path compression), while improving response correctness by 3.3\% and helpfulness by 21.8\%, and drives significant growth in both commercial outcomes (order conversion rate) and user engagement (conversation turns), demonstrating its effectiveness in high-traffic production environments.

\end{abstract}

\ccsdesc[500]{Computing methodologies~Natural language processing}
\keywords{Retrieval-Augmented Generation, Efficient Reasoning, Reinforcement Learning, Mixture-of-Experts, E-commerce Customer Service}

\received{20 February 2007}
\received[revised]{12 March 2009}
\received[accepted]{5 June 2009}

\maketitle


\section{INTRODUCTION}
\label{sec:introduction}

\begin{figure}
    \centering
    \includegraphics[width=0.6\linewidth]{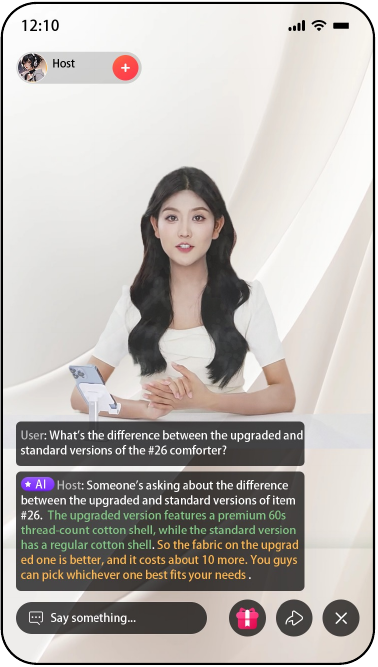}
    \caption{Conversational e-commerce assistance within AI-powered livestreaming.
    \raisebox{-0.2ex}{\includegraphics[height=1.0em]{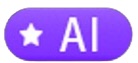}}
    icon indicates the AI's response.
    \textcolor{LightGreen}{Green} text indicates high correctness; \textcolor{LightOrange}{Orange} text indicates high helpfulness.}
    \label{fig:length_reward}
\end{figure}

AI-powered livestreaming leverages conversational agents as real-time e-commerce assistants, forming a challenging Retrieval Augmented Generation (RAG) scenario~\cite{lewis2020retrieval, zhu2023large, gao2023retrieval}. The system must dynamically retrieve relevant product information and generate factually accurate responses under strict latency constraints. Response correctness, helpfulness, and low latency are all critical, directly impacting user engagement and sales performance.

Large Reasoning Models (LRMs), such as OpenAI-o1~\cite{openai2024openaio1card}, Gemini 2.5~\cite{comanici2025gemini25pushingfrontier}, and DeepSeek-R1~\cite{guo2025deepseek}, excel at complex reasoning and generating accurate, nuanced responses. These capabilities are essential for achieving correctness and user-centric helpfulness in AI-powered e-commerce livestreaming. However, their end-to-end response latency can range from tens to hundreds of seconds, due not only to their large-scale parameters but also to lengthy reasoning trajectories. Such delays are unacceptable in interactive settings like live streaming, where audience engagement critically depends on timely responses. Consequently, LRMs face significant challenges in deployment within real-time, low-latency applications.

To reduce the response latency of LRMs for deployment in real-time applications, we propose a practical two-stage optimization framework. The first stage focuses on transferring the reasoning capabilities of a large-scale teacher model to a lightweight student model. Specifically, we distill a 670B teacher model (DeepSeek-R1) into a 30B Mixture-of-Experts (MoE) model Qwen3-30B-A3B\cite{yang2025qwen3technicalreport}, which only activates 3B parameters during inference. To ensure that the student learns high-quality reasoning paths, we employ Rejection Sampling Fine-Tuning (RFT)~\cite{gou2021knowledge, gupta2022compression}, guided by an LLM-based judge that filters out incorrect or unhelpful generations. This yields a distilled model that closely approximates the teacher’s response quality while operating at significantly reduced computational cost. However, due to imitation learning, the student model reproduces the teacher’s lengthy reasoning paths, resulting in end-to-end latency that exceeds industrial requirements despite its compact size. To address this inefficiency, we introduce a second stage that explicitly shortens reasoning trajectories while preserving response quality. Specifically, we employ reinforcement learning with Group Relative Policy Optimization (GRPO)~\cite{deepseek-math} to refine the student model’s reasoning process. The approach uses a multi-objective reward function that jointly optimizes for reasoning brevity, correctness, and helpfulness, encouraging the model to generate concise yet high-quality responses. As a result, it not only reduces inference latency but also enhances output quality.

We validate the effectiveness and generalization of our framework through comprehensive evaluations across industrial, public, and production settings. On the industrial Tblive-E-Commerce QA benchmark, our final model achieves a 30$\times$ reduction in computational cost while surpassing its 670B teacher by +3.3\% in correctness and +21.8\% in helpfulness. On the public multi-hop reasoning dataset MuSiQue, our model exceeds the teacher's performance with +12.3 Exact Match and +10.2 F1-Score, generates 41\% shorter responses, and reduces decoding cost by 95\%. These results highlight the effectiveness of our framework in balancing response quality, reasoning efficiency, and latency, which are key requirements for deployable interactive systems. We deployed the framework in Taobao Live’s\footnote{Taobao Live is one of the largest e-commerce live-streaming platforms in China.} DigitalLive services in July 2025, replacing a 14B dense production model.  It has since driven a 59.0\% increase in order conversion rate\footnote{Order conversion rate is calculated as the number of direct orders divided by the user volume. A ``direct order'' is defined as a viewer purchasing a product that was retrieved by the RAG system during the conversation. The majority of these orders are completed within 15 minutes of the interaction.}, leading to improved GMV performance and a 50.5\% rise in the proportion of multi-turn conversations\footnote{The proportion of multi-turn traffic is calculated as the volume of multi-turn conversations divided by the total traffic volume.}, indicating higher user engagement. 

Our contributions are as follows:

\begin{itemize}
\item \textbf{A two-stage optimization framework for deploying LRMs in real-time settings.} We propose a synergistic pipeline that combines knowledge distillation with reinforcement learning, addressing the challenge of balancing correctness, helpfulness, and low latency in industrial AI assistants.

\item \textbf{Identification and mitigation of inherited reasoning inefficiency in distilled models.} We reveal that standard distillation leads to replication of verbose reasoning paths from teacher to student, and introduce a targeted reinforcement learning stage using GRPO to compress reasoning trajectories while preserving output quality explicitly.

\item \textbf{An efficient MoE-based reasoning model validated on large-scale e-commerce livestreaming.} Our approach achieves significant computational efficiency gains and improved response quality over the teacher model, demonstrating the feasibility of high-performance, low-latency reasoning systems in interactive applications.
\end{itemize}

\section{RELATED WORKS}

\subsection{Reinforcement Learning for LLMs}
Reinforcement learning (RL) addresses critical alignment challenges in LLMs—including hallucinations, harmful content generation, and instruction-following failures—by optimizing models to reflect human preferences through reward modeling \cite{ouyang2022training}. Modern approaches leverage Reinforcement Learning from AI Feedback (RLAIF) to automate preference scoring using LLM judges \cite{lee2023rlaif}, while optimization techniques like Proximal Policy Optimization (PPO) \cite{schulman2017proximal} often focus on singular objectives such as helpfulness or harm reduction. Building on recent innovations, GRPO \cite{guo2025deepseek} improves training efficiency and robustness by adopting relative advantage estimation within grouped candidate responses. 

\subsection{Efficient Reasoning}
Enhancing the reasoning capabilities of LLMs through RL is a prevailing trend \cite{guo2025deepseekr1, openai2025o1}. However, this often induces the ``overthinking'' phenomenon, where models produce computationally expensive and verbose Chain-of-Thought (CoT) paths \cite{chen2024donotthink}. A prominent line of research addresses this by integrating length-based rewards or penalties into the RL framework to encourage conciseness. This approach is heavily influenced by seminal works that analyze the trade-off between reasoning length and stability \cite{yeo2025demystifying}, and implement RL-based pruning strategies, such as O1-Pruner \cite{luo2025o1pruner} and the techniques behind Kimi k1.5 \cite{kimi2025kimi}. RAG-RL \cite{ragrl} uses GRPO to leverage the RAG reasoning capability of the LLM. These methods typically leverage RL to penalize verbosity, a strategy also adopted by numerous other studies to prune or shorten CoT chains while maintaining performance \cite{hou2025thinkprune, ning2025longshort, yuan2025shortrl, xiang2025just}.

\section{PRELIMINARIES}
\label{sec:preliminaries}

\subsection{Problem Definition}
\label{subsec:problem_defination}

\begin{table*}[ht!]
\centering
\caption{Examples of AI model responses with different qualities.}
\label{tab:case-study-reformatted}
\begin{tabularx}{\textwidth}{p{2.5cm} p{6cm} X@{}}
\toprule
\textbf{Scenario} & \textbf{Incorrect Model Response} & \textbf{Correct Model Response} \\
\midrule

\textbf{Context:} A user inquires about a specific product detail based on a livestream with multiple product links. \newline\newline
\textbf{Query:} \textit{``Does the big meat zongzi have fatty meat?"} & 
\textbf{Response:} \textit{``A user asked if the big meat zongzi has fat. The Wufangzhai zongzi gift box in link \#8 contains braised pork belly zongzi, and the ingredient list clearly states it's made with three-parts-fat, two-parts-lean pork belly. You can check the details page of link \#8."} \newline\newline
\textbf{Evaluation: \textcolor{red}{Incorrect}}. The model hallucinates information. It incorrectly attributes the ``braised pork belly" detail, which belongs to product \#9, to product \#8 (a gift box). This misrepresentation provides the user with factually wrong information about a specific product, undermining trust and accuracy. & 
\textbf{Response:} \textit{``A user is asking if the big meat zongzi has fat. According to the product information, the braised pork belly zongzi in link \#9 and the pork zongzi in link \#5 both contain pork belly or lean meat, where pork belly is a mix of fat and lean. The ingredient list for the gift box in link \#8 only mentions `pork' without specifying fat content. You can refer to the product images for more details."} \newline\newline
\textbf{Evaluation: \textcolor{blue}{Correct}}. The model demonstrates superior accuracy by correctly synthesizing information from multiple sources. It accurately identifies which products explicitly mention fatty meat (links \#9 and \#5) and correctly points out the ambiguity in the description for product \#8. This response is precise, comprehensive, and factually sound. \\
\midrule
\textbf{Scenario} & \textbf{Unhelpful Model Response} & \textbf{Helpful Model Response} \\
\midrule

\textbf{Context:} A user asks for a product that is not currently available in the provided product list. \newline\newline
\textbf{Query:} \textit{``Do you have a screen protector for iPhone 11?"} & 
\textbf{Response:} \textit{``A user is asking about a screen protector for iPhone 11. We currently do not have a protector for the iPhone 11 in our livestream. You can try consulting customer service to see if it's available through other channels."} \newline\newline
\textbf{Evaluation: \textcolor{red}{Unhelpful}}. The model provides a minimal, dismissive answer. While correct in stating the product is unavailable, it offers no alternative value. By immediately deflecting the user to customer service, it fails to engage them or leverage the available product information to provide any assistance. & 
\textbf{Response:} \textit{``A user is asking for an iPhone 11 screen protector. Currently, link \#13 is for models like iPhone 15 and 14 Pro Max, and link \#12 is for models like 12 Pro and 13 Pro. Link \#1 is for a phone case. We don't have a protector for the iPhone 11 in the product list, it might be out of stock or not listed. You can double-check the links or ask customer service about other models."} \newline\newline
\textbf{Evaluation: \textcolor{blue}{Helpful}}. Despite the requested item being unavailable, the model provides a highly helpful response. It proactively informs the user about which products \textit{are} available (protectors for iPhone 12, 13, 14, 15 series) and their corresponding links. This action adds value beyond the direct query, helps manage user expectations, and potentially redirects the user's interest to a relevant, available product. \\

\bottomrule
\end{tabularx}
\end{table*}

We define a conversational commerce assistance task where, given a viewer question about products, the system must generate helpful and accurate responses. 
Formally, for each interaction at turn $t$, the inputs consist of \textbf{Retrieved Product Documents ($D_t$)}, which include relevant product specifications and attributes retrieved from the catalog; the \textbf{Audience Question ($Q_t$)}, representing the current natural language query; and the \textbf{Conversation History ($H_t = \{(Q_{1:t-1}, R_{1:t-1})\}$)}, which provides the preceding dialogue context.
Based on these inputs, the system outputs a \textbf{Thought Process ($T_t$)} detailing the step-by-step reasoning, and a \textbf{Final Response ($R_t$)} as the natural language answer to the viewer's query.

\subsection{Evaluation Framework}
\label{subsec:evaluation_framework}
Responses are evaluated along two dimensions using \textsc{LLM-as-a-Judge}:
\paragraph{\textbf{Correctness (C)}}
Measures factual alignment with product knowledge:
\[
C(D_t, Q_t, H_t, R_t) = 
\begin{cases} 
1 & \text{if } R_t \text{ accurately reflects } D_t, \\
0 & \text{otherwise.}
\end{cases}
\]
A response is deemed correct if and only if all of its claims are verifiable against the provided documents ($D_t$), it maintains logical consistency with the conversation history ($H_t$), and it serves as a direct and pertinent answer to the current query ($Q_t$). This ensures the response is not only factually sound but also avoids being non-responsive or ``off-topic."

\paragraph{\textbf{Helpfulness (H)}}
While correctness is a fundamental requirement, optimizing for correctness alone can yield undesirable model behavior. For instance, a model might become overly cautious, generating minimal responses to minimize the risk of hallucination. Although factually correct, such terse answers deliver a poor user experience by failing to be genuinely assistive.

To counteract this tendency, we introduce Helpfulness as a complementary metric. It incentivizes the model to provide richer, more valuable information, striking a balance between accuracy and utility:
\[
H(Q_t, H_t, R_t) = 
\begin{cases} 
1 & \text{if } R_t \text{ proactively assists the user } Q_t,  \\
0 & \text{otherwise.}
\end{cases}
\]
A response is considered helpful if, in addition to correctly answering the query, it proactively offers supplementary, relevant information from the documents that anticipates and addresses the user's implicit needs within the conversational context ($H_t$).

\paragraph{\textbf{Validating the LLM Judge}}
To ensure the reliability of our \textsc{LLM-as-a-Judge}, we validated its performance against human judgment. We began by manually annotating a corpus of several hundred of responses for both Correctness and Helpfulness. These were treated as two independent binary classification tasks. Each response was labeled by two annotators, and to maintain high data quality, instances with conflicting labels were discarded. We then balanced the positive and negative samples for each task to a 1:1 ratio, creating two distinct benchmark datasets. Through iterative refinement of the judge prompt, our LLM judge achieved an accuracy of \textbf{83\%} on the Correctness benchmark and \textbf{84\%} on the Helpfulness benchmark, demonstrating strong alignment with human evaluation.

\vspace{0.5em}
\begin{figure*}[t]
    \centering
    \includegraphics[width=0.95\linewidth]{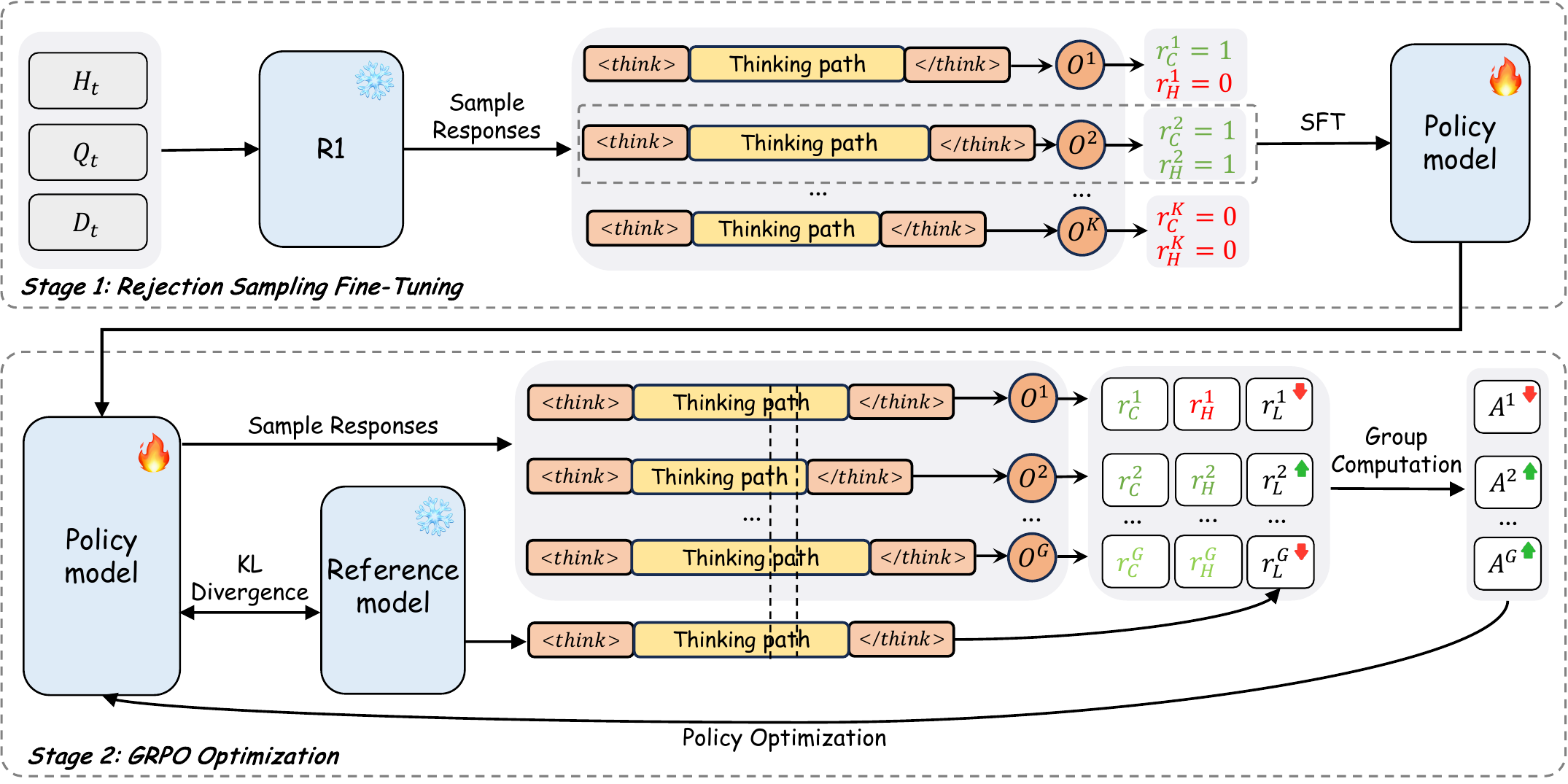} 
    \vskip -0.1in
    \caption{The proposed two-stage methodology consists of: (1) an initial knowledge distillation stage using RFT, followed by (2) a second stage dedicated to optimizing the efficiency of the reasoning path via GRPO.}
    \vskip -0.1in
    \hfill
\end{figure*}

\section{METHODOLOGY}
\label{sec:methodology}
Our two-stage methodology first distills e-commerce reasoning ability into a lightweight model via RFT. Subsequently, we employ GRPO to compress the model's reasoning path, optimizing for efficiency while maintaining response quality.


\subsection{E-Commerce RAG Reasoning Ability Distillation based on RFT}
\label{subsec:distill}

To transfer capabilities from the 670B reasoning MoE teacher model (37B activated parameters) to the lightweight student model, we employ a three-stage distillation pipeline formalized as:

\paragraph{\textbf{Teacher-based Data Generation}}
Given an unlabeled e-commerce corpus $\mathcal{U} = \{(D^{(i)}_t, Q^{(i)}_t, H^{(i)}_t)\}_{i=1}^N$, we use the teacher model $P_{\theta_{\text{teacher}}}$ with temperature-scaled decoding to sample $k$ distinct candidate trajectories $\{(T^{(i,j)}_t, R^{(i,j)}_t)\}_{j=1}^k$ for each instance $i$. This process results in a candidate set of trajectories, where each full data point is a tuple containing the original input and a generated output.
    
\paragraph{\textbf{Rejection Sampling by LLM Judge Filtering}}
To ensure the quality of our distillation data, we introduce a filtering step that retains only the most valuable trajectories. A trajectory is considered valuable if it is evaluated as both correct and helpful. We employ an LLM-as-Judge, implemented via DeepSeek-R1, to perform this evaluation based on the metrics defined in \S\ref{sec:preliminaries}.

Specifically, for each input instance $i$, we iterate through its $k$ generated trajectories. If at least one trajectory $(T_t^{(i,j)}, R_t^{(i,j)})$ satisfies both conditions, we select one such valid trajectory to be included in our final distillation dataset, denoted as $\mathcal{D}_{\text{distill}}$. If none of the $k$ trajectories for a given input pass the filters, the instance is discarded entirely. The final dataset is constructed as follows:
\[
\mathcal{D}_{\text{distill}} \subseteq \bigcup_{i=1}^N \bigcup_{j=1}^k \left\{ 
    \tau^{(i,j)} \mid C(\tau^{(i,j)}) = 1 \land H(\tau^{(i,j)}) = 1
\right\},
\]
where $\tau^{(i,j)} = (D_t^{(i)}, Q_t^{(i)}, H_t^{(i)}, T_t^{(i,j)}, R_t^{(i,j)})$, and the judgment functions for correctness $C(\cdot)$ and helpfulness $H(\cdot)$ are applied accordingly. The subset notation ($\subseteq$) reflects our strategy of selecting at most one trajectory per original input instance.

\paragraph{\textbf{Student MoE Model Fine-tuning}} 
Since our student model $\theta^{\text{student}}$ is a Mixture of Experts (MoE) model, we train it by minimizing a composite loss function. This objective includes the primary sequence-to-sequence task loss and an auxiliary loss for load balancing. The total optimization objective is:
\[
\mathcal{L}_{\text{total}} = \mathcal{L}_{\text{SFT}} + \lambda_{\text{aux}} \cdot \mathcal{L}_{\text{aux}.}
\]

The main task loss $\mathcal{L}_{\text{SFT}}$ trains the model to reconstruct teacher outputs:
\[
\mathcal{L}_{\text{SFT}} = -\mathbb{E}_{(D,Q,H,T,R) \sim \mathcal{D}_{\text{distill}}} \left[ \sum_{i=1}^{L_y} \log P_{\theta^{\text{student}}} \left( y_i \mid \mathbf{x}, y_{<i} \right) \right],
\]
where the input $\mathbf{x} = [D; Q; H]$ is the concatenated context, and the target sequence $y = [T; R]$ encodes the thought $T$ and response $R$. The term $\mathcal{L}_{\text{aux}}$ represents the MoE auxiliary loss, which promotes balanced routing of tokens across experts. Its contribution is scaled by the hyperparameter $\lambda_{\text{aux}}$, ensures stable training and effective expert specialization.

\subsection{Jointly Optimizing Reasoning Efficiency, Helpfulness, and Correctness with RL}
\label{subsec:rl}

While RFT (as detailed in \S\ref{subsec:distill}) transfers knowledge, it also inherits the teacher's verbose reasoning. We address this using RL to optimize for reasoning efficiency, correctness, and helpfulness.

\paragraph{\textbf{RL Problem Formulation}}
We frame response generation as a Markov Decision Process (MDP). The state $s_t$ is the concatenated input context $[D_t; Q_t; H_t]$, and the action $a_t$ is a token sequence $[T_t; R_t]$ generated autoregressively by the policy $\pi_\theta$, a student model with parameters $\theta$. An episode consists of the complete generation of an action $a_t$ and terminates upon producing an end-of-sequence (EOS) token.

\paragraph{\textbf{Reward Design}}
\label{para:reward}

To enhance the model's reasoning efficiency while ensuring the quality of its responses, we propose a multi-objective RL optimization framework. This framework is implemented through a set of meticulously designed reward functions. First, we introduce a length-based reward function to optimize for computational efficiency. Concurrently, we formulate correctness and helpfulness reward functions to constrain the model, thereby guaranteeing that the pursuit of efficiency does not compromise the quality of its outputs.

\textbf{Length Reward ($r_{\mathrm{length}}$)}: 
Optimizes reasoning efficiency relative to reference model outputs:
\begin{align*}
L_{\mathrm{upper}} &= \lambda_{\mathrm{upper}} L_{\mathrm{ref}}, \quad 
L_{\mathrm{lower}} = \lambda_{\mathrm{lower}} L_{\mathrm{ref},} \\
d &= \begin{cases} 
\max(L_{\mathrm{lower}} - L_{\mathrm{policy}}, 0) & L_{\mathrm{policy}} < L_{\mathrm{lower},} \\
\max(L_{\mathrm{policy}} - L_{\mathrm{upper}}, 0) & L_{\mathrm{policy}} > L_{\mathrm{upper},} \\
0 & \text{otherwise,}
\end{cases} \\
r_{\mathrm{length}} &= \max\left(0, 1 - \frac{d}{\epsilon}\right).
\end{align*}

The length reward $r_{\mathrm{length}}$, is calculated based on how the length of a policy-generated sequence $L_{\mathrm{policy}}$, compares to an acceptable length range. This range is determined by the length of a reference sequence $L_{\mathrm{ref}})$. The reference sequences are generated from the reference model, which in our case is the student model after the RFT stage. Specifically, we define an upper bound $L_{\mathrm{upper}} = \lambda_{\mathrm{upper}} L_{\mathrm{ref}}$ and a lower bound $L_{\mathrm{lower}} = \lambda_{\mathrm{lower}} L_{\mathrm{ref}}$. The coefficients $\lambda_{\mathrm{upper}}$ and $\lambda_{\mathrm{lower}}$ control the desired compression ratio and prevent over-compression, respectively. The deviation $d$ measures how far $L_{\mathrm{policy}}$ is outside the acceptable interval $[L_{\mathrm{lower}}, L_{\mathrm{upper}}]$. Finally, the reward $r_{\mathrm{length}}$ is computed by penalizing this deviation, with a hyperparameter $\epsilon$ serving as a tolerance margin that dictates how sharply the reward decreases with the deviation.

\begin{figure}[t]
    \centering
    \includegraphics[width=0.9\linewidth]{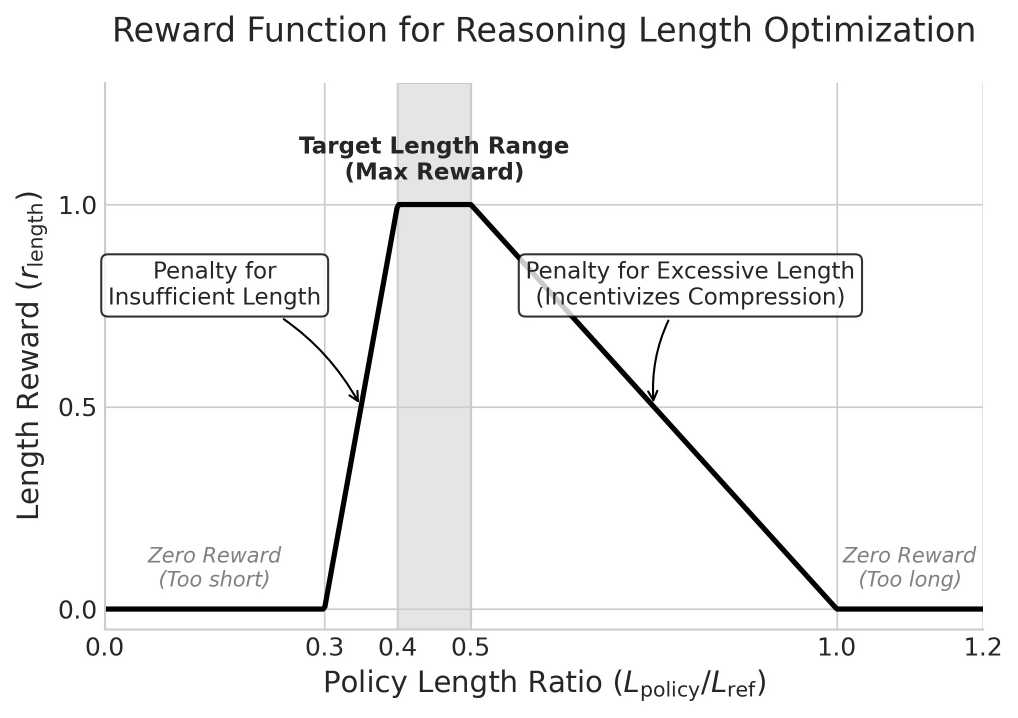}
    \vskip -0.1in
    \caption{Reward Function for Reasoning Length Optimization}
    \vskip -0.1in
    \label{fig:length_reward}
\end{figure}

\textbf{Correctness Reward ($r_{\mathrm{correct}}$)}: 
Reduces hallucinations by verifying response fidelity to product documents:
\begin{equation*}
r_{\mathrm{correct}} = C(D_t, Q_t, H_t, R_t) \in \{0,1\},
\end{equation*}
where 1 indicates $R_t$ is fully consistent with $D_t$ and answers $Q_t$ without fabrication.

\textbf{Helpfulness Reward ($r_{\mathrm{helpful}}$)}: 
Prevents over-conservatism by assessing practical utility:
\[
r_{\mathrm{helpful}} = H(Q_t, H_t, R_t) \in \{0,1\},
\]
where 1 denotes $R_t$ provides actionable shopping guidance addressing implicit needs in $H_t$.

\paragraph{\textbf{Group Relative Advantage Calculation}}
To derive a normalized signal for policy optimization, we first compute a composite reward $r_j$ for each of the $K$ responses in a batch (Eq.~\eqref{eq:composite_reward_condensed}). This score is a weighted sum of its correctness, helpfulness, and length-based rewards. The advantage score $A_j$ is then calculated by normalizing $r_j$ using the mean $\mu$ and standard deviation $\sigma$ of the rewards across the batch, as shown in Eq.~\eqref{eq:advantage_calc_condensed}. This normalization transforms the raw rewards into standardized Z-scores, providing a stable signal for policy updates. A small constant $\epsilon_{\mathrm{std}}$ (e.g., $10^{-8}$) ensures numerical stability.

\begin{align}
    r_j &=  w_c r_j^{\mathrm{correct}} + w_h r_j^{\mathrm{helpful}} + w_l r_j^{\mathrm{length}}, \label{eq:composite_reward_condensed} \\
    A_j &= \frac{r_j - \mu}{\sigma + \epsilon_{\mathrm{std}}}, \quad \text{where } \mu, \sigma = \text{mean, stddev}\left( \{r_i\}_{i=1}^K \right) .\label{eq:advantage_calc_condensed}
\end{align}
Intuitively, this mechanism rewards responses that perform better than average within a mini-batch, making training more stable and less sensitive to absolute reward scales.

\paragraph{\textbf{Policy Optimization}}
The GRPO objective maximizes advantage-regularized returns:
\begin{align*}
\mathcal{L}^{\mathrm{GRPO}}(\theta) = \frac{1}{K} \sum_{j=1}^K \Bigg[ & \min\left( \frac{\pi_\theta(a_j|s_j)}{\pi_{\theta_{\text{old}}}(a_j|s_j)} A_j, \right. \\
& \left. \text{clip}\left(\frac{\pi_\theta(a_j|s_j)}{\pi_{\theta_{\text{old}}}(a_j|s_j)}, 1-\eta, 1+\eta\right) A_j \right) \\
& - \beta D_{\text{KL}}\left( \pi_{\theta_{\text{old}}} \| \pi_\theta \right) \Bigg],
\end{align*}
where $\eta=0.2$ is the clip range, $\beta=0.01$ controls KL penalty, and $s_j = (D_t, Q_t, H_t)$, $a_j = (T_t,  R_t)$. We choose GRPO over standard PPO due to its improved training efficiency and stability, particularly in scenarios with complex, multi-objective rewards.

\section{EXPERIMENTAL SETTINGS}

\subsection{Datasets}
We evaluate our approach on two distinct datasets, each containing 8,000 training and 400 test samples.

\paragraph{\textbf{Tblive-E-Commerce QA}} This proprietary dataset is sourced from live shopping streams on our e-commerce platform. Each sample comprises a conversation history, a viewer's real-time question, and relevant structured product information that was dynamically retrieved based on user intent. This dataset encapsulates key industrial challenges, such as reasoning over noisy conversational context and grounding responses in structured merchandise data.

\paragraph{\textbf{MuSiQue}} As a public benchmark for multi-hop question answering, MuSiQue tests a model's complex reasoning ability. Each question is designed to be unanswerable from any single document, requiring the model to synthesize information across multiple paragraphs. We utilize the distractor setting, where each question is paired with approximately 10 documents, most of which are irrelevant distractors. This setup rigorously evaluates the model's capacity to identify and connect relevant facts from noisy, knowledge-intensive contexts, complementing our industrial dataset.

\paragraph{\textbf{Evaluation}} Evaluation protocols differ across the test sets: On Tblive-E-Commerce QA, responses are assessed using the \textit{LLM-as-a-Judge} framework from \ref{subsec:evaluation_framework} for both correctness (C) and helpfulness (H). For MuSiQue, we follow standard evaluation practice using Exact Match (EM) and F1 scores to measure factual accuracy.


\begin{table}[t]
  \centering
  \small 
  \caption{Hyperparameters for Distillation and GRPO.} 
  \vskip -0.1in
  \label{tab:hyper_parameters}
  \begin{tabular}{@{}ll@{\hspace{1.5em}}ll@{}}
    \toprule
    \multicolumn{2}{c}{\textbf{Distillation Hyperparameters}} & \multicolumn{2}{c}{\textbf{GRPO Hyperparameters}} \\
    \cmidrule(r){1-2} \cmidrule(l){3-4} 
    \textbf{Parameter} & \textbf{Value} & \textbf{Parameter} & \textbf{Value} \\
    \midrule
    Teacher model   & DeepSeek-R1 & Rollouts per input ($K$) & 8 \\
    Base model   & Qwen3-30B-A3B   & Correctness weight ($w_c$) & 0.4 \\
    LLM judge model & DeepSeek-R1           & Helpfulness weight ($w_h$) & 0.3 \\
    Temperature ($\tau$) & 0.7              & Length weight ($w_l$) & 0.3 \\
    Samples ($k$) & 4             & Upper length ratio ($\lambda_{\mathrm{upper}}$) & 0.5 \\
    Batch size      & 2                     & Lower length ratio ($\lambda_{\mathrm{lower}}$) & 0.4 \\
    Learning rate   & 1.5e-6                  & Length tolerance ($\epsilon$) & 0.1 \\
    $\lambda_{\mathrm{aux}}$    & 0.01                      & Learning rate & 2.5e-6 \\
                    &                       & Batch size    & 8 \\
    \bottomrule
  \end{tabular}
  \vskip -0.15in
\end{table}

\subsection{Experimental Setup}
\label{subsec:experimental_setup}

We implement our two-stage training framework using specialized infrastructure for each phase. The distillation phase employs our internal Magatron-based framework for RFT, processing the filtered $\mathcal{D}_{\text{filtered}}$ dataset through large-scale distributed training. For the RL phase, we utilize the ROLL framework \cite{roll2025alibaba} to implement the GRPO described in \S\ref{subsec:rl}. Reward configurations are adapted per dataset: For Tblive-E-Commerce QA, we apply the composite reward with components defined in \ref{para:reward} (correctness, helpfulness, and length rewards). For MuSiQue, we maintain the length reward but substitute correctness assessment with EM as a binary reward ($r_{\mathrm{correct}} = \text{EM score}$), while omitting helpfulness due to the dataset's non-conversational nature. The experimental hyperparameters are detailed in Table~\ref{tab:hyper_parameters}.

\section{EXPERIMENTS}

\begin{table*}[!ht]
\centering
\small
\caption{Experimental Results on Tblive-E-Commerce QA and MuSiQue Datasets. Reported metrics include correctness (Correct.), helpfulness (Helpful.), tokens per response (TPR), and decoding computational cost (D. TFLOPs).}
\label{tab:results}
\begin{tabular}{lcc|cccc|cccc}
\toprule
\multirow{2}{*}{Model} & \multicolumn{2}{c|}{Parameters (B)} & \multicolumn{4}{c|}{Tblive-E-Commerce QA} & \multicolumn{4}{c}{MuSiQue} \\
\cmidrule(lr){2-3} \cmidrule(lr){4-7} \cmidrule(lr){8-11}
& Total & Active & Correct. & Helpful. & TPR & D. TFLOPs & EM & F1 & TPR & D. TFLOPs \\
\midrule
DeepSeek-R1 & 670 & 37 & 89.0 & 75.0 & 421 & 31.15 & 41.2 & 57.0 & 807 & 59.72 \\
Qwen3-235B-A22B & 235 & 22 & 78.8 & 71.5 & 387 & 17.03 & 38.5 & 50.0 & 881 & 38.76 \\
Qwen3-32B & 32 & 32 & 76.5 & 76.3 & 368 & 23.55 & 37.5 & 52.5 & 769 & 49.22 \\
Qwen3-30B-A3B & 30 & 3 & 67.3 & 56.5 & 396 & 2.38 & 35.0 & 46.8 & 995 & 5.97 \\
DeepSeek-R1-Distilled-Llama-8B & 8 & 8 & 20.8 & 34.5 & 332 & 5.31 & 17.5 & 29.0 & 625 & 10.00 \\

\midrule

DeepSeek-R1-Distilled-Llama-8B + RFT & 8 & 8 & 75.8 & 77.8 & 334 & 5.34 & 38.3 & 50.0 & 734 & 11.74 \\
DeepSeek-R1-Distilled-Llama-8B + RFT + RL & 8 & 8 & 86.3 & 93.8 & 152 & 2.43 & 46.0 & 58.4 & 412 & 6.59 \\
Qwen3-30B-A3B + RFT & 30 & 3 & 81.3 & 77.0 & 353 & 2.12 & 46.8 & 58.7 & 762 & 4.57 \\
Qwen3-30B-A3B + RL & 30 & 3 & 87.5 & 90.0 & $\textbf{152}$ & $\textbf{0.91}$ & 50.2 & 65.8 & $\textbf{450}$ & $\textbf{2.70}$ \\
Qwen3-30B-A3B + RFT + RL (LiveThinking) & 30 & 3 & $\textbf{92.3}$ & $\textbf{96.8}$ & 153 & 0.92 & $\textbf{53.5}$ & $\textbf{67.2}$ & 473 & 2.84 \\
\bottomrule
\end{tabular}
\vspace{0.2cm}
\end{table*}

\subsection{Main Results}

\label{subsec:main_results_and_analysis}


Table~\ref{tab:results} presents the performance of our proposed framework on both the industrial \textit{Tblive-E-Commerce QA} and public \textit{MuSiQue} benchmarks. Our final model, \textit{LiveThinking} (Qwen3-30B-A3B + RFT + RL), achieves state-of-the-art results across all metrics while operating at industrial-scale efficiency.

On the Tblive-E-Commerce QA dataset, \textit{LiveThinking} attains \textbf{92.3\% correctness} and \textbf{96.8\% helpfulness}, surpassing its 670B teacher (\textit{DeepSeek-R1}: 89.0\%, 75.0\%) and all baselines by significant margins. Despite being 22$\times$ smaller in total parameters and activating only 3B parameters during inference, it reduces decoding computation to just \textbf{0.92 TFLOPs}, which is over 30$\times$ less than the teacher, while generating responses that are more than \textbf{2.7$\times$ shorter}. This demonstrates that careful optimization enables small models to exceed larger teachers in both quality and efficiency. More case studies are detailed in Appendix~\ref{app:case_study}.

Consistent gains are observed on MuSiQue: \textit{LiveThinking} achieves \textbf{53.5 EM} and \textbf{67.2 F1}, outperforming the 670B teacher (41.2 EM, 57.0 F1) and setting a new state-of-the-art among models with comparable or even much higher computational cost. Response length is reduced from 807 to 473 tokens, confirming the generalizability of our approach beyond domain-specific data.

Notably, our framework also achieves strong performance on a dense 8B Llama backbone: the \textit{DeepSeek-R1-Distilled-Llama-8B + RFT + RL} variant improves correctness from 20.8\% to 86.3\% and helpfulness from 34.5\% to 93.8\%, while reducing TPR from 332 to 152. This demonstrates the generality of our two-stage optimization approach across different model architectures.

\subsection{Ablation Study}
\label{subsec:ablation}
To evaluate the role of each stage, we conduct an ablation study on the Qwen3-30B-A3B backbone (Table~\ref{tab:results}).
As expected, distillation with RFT improves correctness from 67.3\% to 81.3\% and helpfulness from 56.5\% to 77.0\%, confirming its ability to transfer high-quality reasoning behaviors. However, the student inherits the teacher’s verbose reasoning patterns, resulting in only a slight reduction in response length, i.e., 396 vs. 353 TPR. RL alone achieves higher correctness (87.5\%) and helpfulness (90.0\%) with significantly shorter responses (TPR: 152), demonstrating that policy optimization can simultaneously improve both quality and efficiency. Crucially, when RFT is used to initialize the model before RL, the final performance further increases to \textbf{92.3\% correctness} and \textbf{96.8\% helpfulness}. This significant gain suggests that the RFT stage provides a stronger behavioral prior by aligning the model with correct multi-step reasoning, which the RL stage can then refine more effectively under latency constraints. Thus, the two stages are complementary: RFT establishes a solid foundation of reasoning capability, while RL compresses and optimizes it for deployment. The synergy enables performance beyond what either method can achieve independently.



\subsection{Efficacy of Rejection Sampling Fine-Tuning}
\label{sec:ablation_rft_sft}

\begin{table}[!htbp]
\centering
\caption{Comparison between Rejection Sampling Fine-Tuning (RFT) and Supervised Fine-Tuning (SFT) for knowledge distillation. Both models are fine-tuned from the Qwen3-30B-A3B baseline.}
\label{tab:ablation_rft_sft}
\begin{tabular}{lcccc}
\toprule
\textbf{Method} & \textbf{Correctness} & \textbf{Helpfulness} \\
\midrule
SFT & 81.0 & 72.0 \\
RFT & \textbf{81.3} & \textbf{77.0} \\
\bottomrule
\end{tabular}
\end{table}

To validate the effectiveness of RFT in the first stage of our framework, we compare it with standard SFT, where the student learns from the same teacher model but without data filtering. As shown in Table~\ref{tab:ablation_rft_sft}, RFT outperforms SFT on both correctness and helpfulness, with a substantial gain in the latter (+5.0\%). We attribute this to the nature of the two metrics under open-ended e-commerce QA. Correctness is well-grounded in provided product information, allowing SFT to preserve it effectively through imitation. In contrast, helpfulness varies significantly across reasoning paths and depends on nuanced aspects such as clarity, structure, and user intent alignment—qualities that are not uniformly present in the teacher’s outputs. By filtering trajectories using an LLM judge, RFT ensures the student learns only from high-quality, well-articulated responses. These results demonstrate that careful curation of training data—rather than indiscriminate imitation—is essential for transferring complex, user-centric reasoning patterns in real-world applications.

\subsection{Impact of Target Reasoning Length}
\label{sec:ablation_length}

\begin{figure}[t]
    \centering
    \includegraphics[width=0.9\linewidth]{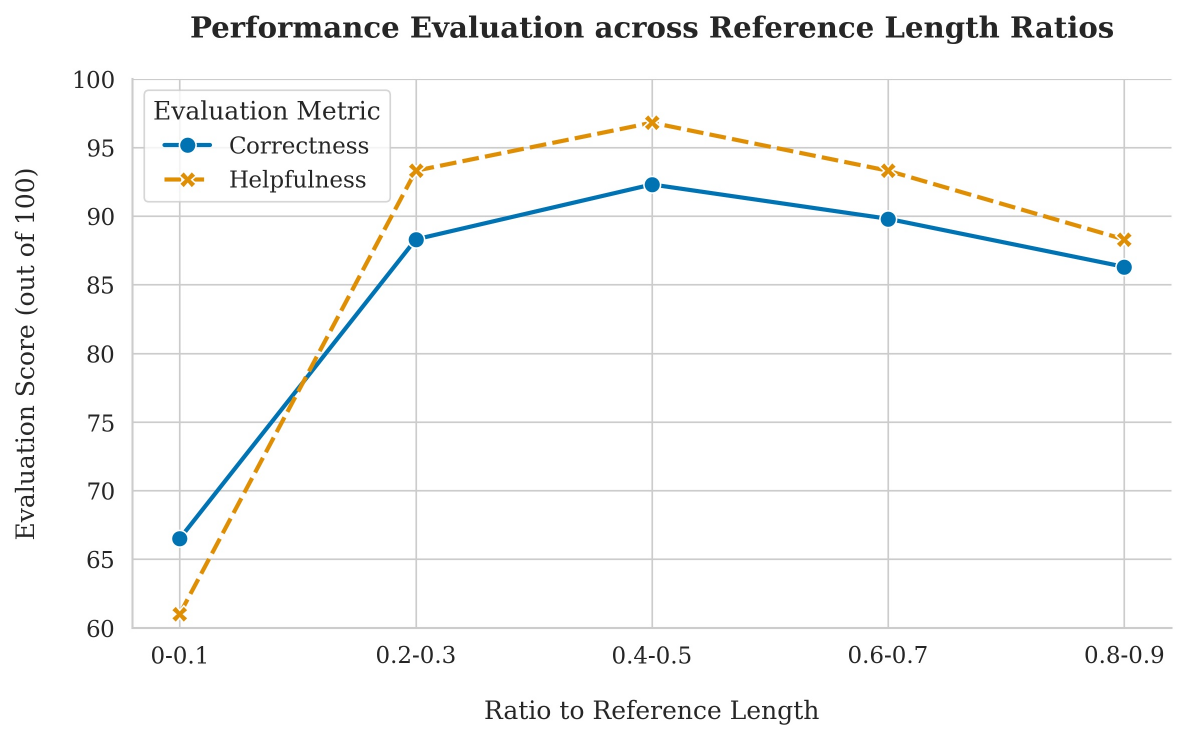}
    \caption{Performance Evaluation across Length Ratio}
    \label{fig:target_length}
    \vskip -0.15in
\end{figure}


A key aspect of our RL-based compression stage is balancing reasoning brevity and response quality. To investigate this trade-off, we train variants of the model with different target reasoning lengths, controlled by adjusting the coefficients $\lambda_{upper}$ and $\lambda_{lower}$ in the reward function. Figure \ref{fig:target_length} presents the results of this sensitivity analysis. The results reveal a clear trend: when the brevity constraint is too strong, the model generates excessively short reasoning paths that lack sufficient intermediate steps, leading to drops in correctness and helpfulness. As the constraint is relaxed, performance improves, peaking at an intermediate setting where the model maintains enough internal structure to reason effectively while remaining concise. However, when the brevity constraint is too weak, the model generates unnecessarily long reasoning paths that often repeat key premises or introduce irrelevant thoughts, leading to performance degradation. Our goal is to identify an optimal operating point that significantly reduces the average number of tokens per response—thereby lowering latency—without compromising correctness and helpfulness scores. This experiment validates the specific brevity target used for our final model.

\subsection{Ablation Study of the Multi-Objective Reward Function}
\label{sec:ablation_rewards}

\begin{table}[t]
\centering
\caption{Ablation study on the components of the multi-objective reward function. All models are trained from Qwen3-30B-A3B + RFT model.}
\vskip -0.1in
\label{tab:ablation_rewards}
\begin{tabular}{lccc}
\toprule
\textbf{Rewards} & \textbf{Correctness} & \textbf{Helpfulness} & \textbf{TPR} \\
\midrule
$R_{\text{correct}}$ + $R_{\text{helpful}}$ & 91.5 & 95.0 & 349 \\
$R_{\text{correct}}$ + $R_{\text{length}}$   & 87.3 & 72.3 & 145 \\
$R_{\text{helpful}}$ + $R_{\text{length}}$   & 79.3 & \textbf{96.8} & \textbf{152} \\
\midrule
Full & \textbf{92.3} & \textbf{96.8} & 153 \\
\bottomrule
\end{tabular}
\end{table}

We conduct an ablation study to demonstrate the indispensability of each component in our multi-objective reward function: correctness ($R_{\text{correct}}$), helpfulness ($R_{\text{helpful}}$), and brevity ($R_{\text{length}}$). As shown in Table~\ref{tab:ablation_rewards}, removing any single component leads to a notable degradation in the corresponding metric.

Notably, including the brevity reward not only reduces latency but also enhances both correctness and helpfulness. This suggests that optimizing for brevity encourages the model to discover more efficient and accurate reasoning paths. Furthermore, when helpfulness is not optimized (e.g., $R_{\text{correct}} + R_{\text{length}}$), the model becomes overly conservative to reduce hallucinations, sacrificing helpfulness. In contrast, simultaneously optimizing for helpfulness and correctness forces the model to genuinely improve its RAG reasoning capabilities to mitigate hallucinations, rather than merely avoiding them, resulting in a significant boost in correctness (91.5 vs. 87.3). These findings confirm that all three components are essential for holistically addressing the correctness-helpfulness-latency trilemma.

\subsection{Architectural Analysis: Efficacy of MoE}
\label{sec:ablation_moe}

\begin{figure}
    \centering
    \includegraphics[width=1.\linewidth]{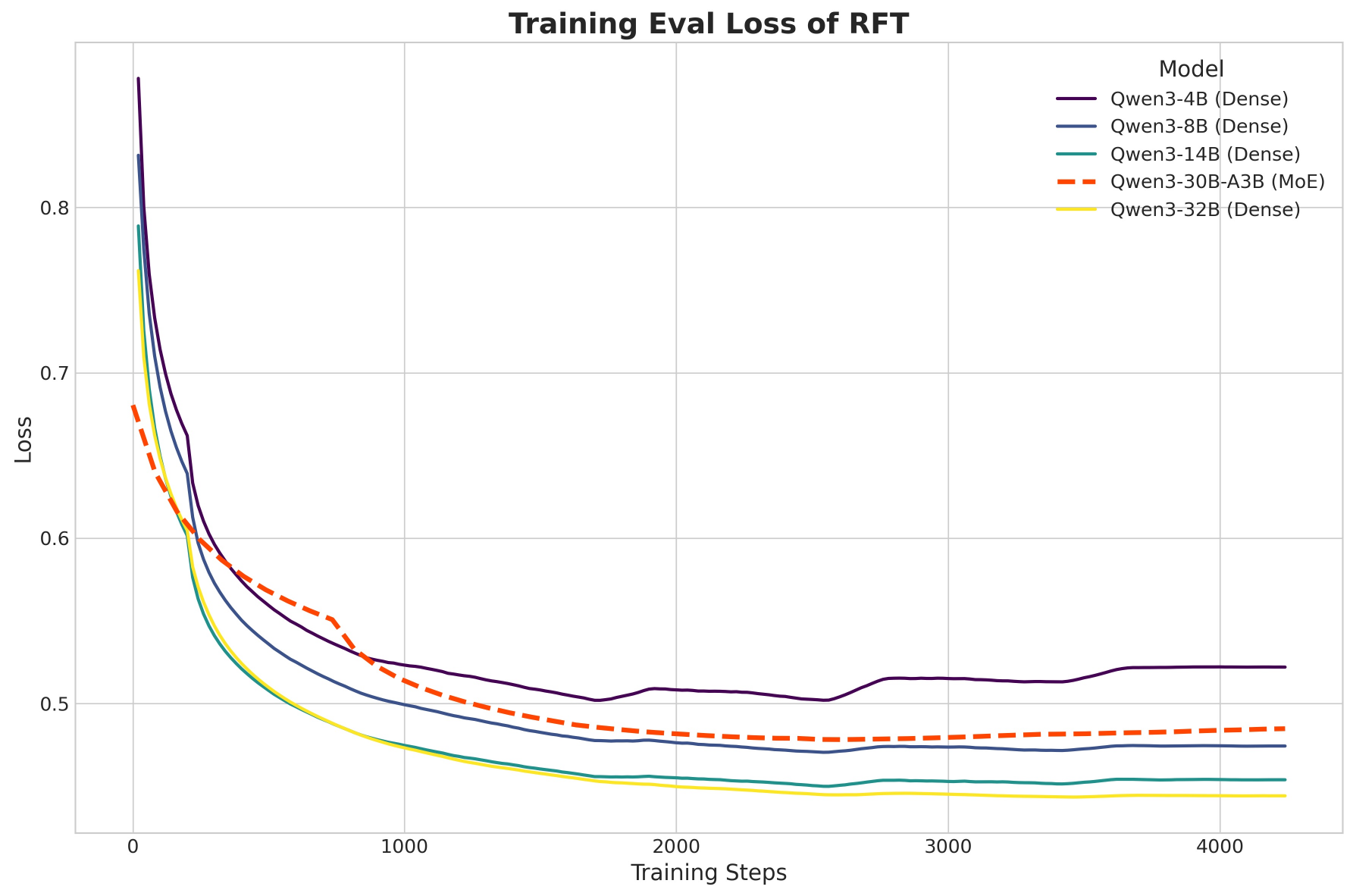}
    \vskip -0.05in
    \caption{RFT Training Eval Loss Curve of Different Models}
    \label{fig:length_reward}
    \vskip -0.05in
\end{figure}

To justify our choice of a MoE architecture, we compare its perform-ance-to-cost ratio against traditional dense models. We benchmark our final model, built on Qwen3-30B-A3B (30B total, 3B active), against a range of dense models spanning from 4B to 32B parameters. Figure \ref{fig:length_reward} illustrates that the MoE architecture achieves performance approaching that of the large dense model (about 7.5B) in our scenario while requiring a fraction of the computational cost during inference. It also vastly outperforms the small dense model. This confirms that MoE is the most efficient architectural choice for our latency-critical application.


\section{CONCLUSION}

In this paper, we introduce LiveThinking, a two-stage training framework that adapts LRMs to real-time applications by balancing quality and latency. Our approach first leverages RFT to distill reasoning ability from a 670B teacher into a lightweight 30B student model. Recognizing that standard distillation inherits the teacher’s verbose reasoning trajectories, we introduce a second stage that applies GRPO with a multi-objective reward function to explicitly compress reasoning paths while further enhancing correctness and helpfulness. Extensive evaluations demonstrate that LiveThinking achieves a remarkable 30-fold reduction in computational cost, while simultaneously outperforming its much larger teacher on correctness and helpfulness. Deployed at scale on Taobao Live, LiveThinking demonstrates the feasibility of deploying low-latency, high-performance reasoning models in interactive e-commerce applications. This work provides an effective paradigm for deploying powerful yet efficient conversational AI in real-world, time-sensitive settings.

\section{ETHICAL USE OF DATA}
This research adheres to ACM's policies on research involving human subjects. Our proprietary Tblive-E-Commerce QA dataset was fully anonymized by removing all Personally Identifiable Information (PII) to protect user privacy; the MuSiQue dataset is a public benchmark. Ethical risks such as misinformation are directly mitigated by our methodology: the Correctness reward minimizes hallucinations, while the Helpfulness reward promotes genuine user value. The research was conducted in accordance with our organization's internal data governance and ethics review policies.

\bibliographystyle{ACM-Reference-Format}
\bibliography{references}

\appendix

\section{LLM Usage Description}
We used a generative AI to assist with Chinese-to-English translation, language polishing, and LaTeX table formatting. The authors have reviewed all AI-generated content and take full responsibility for the final paper.

\begin{figure}[htbp]
    \centering
    \includegraphics[width=0.7\linewidth]{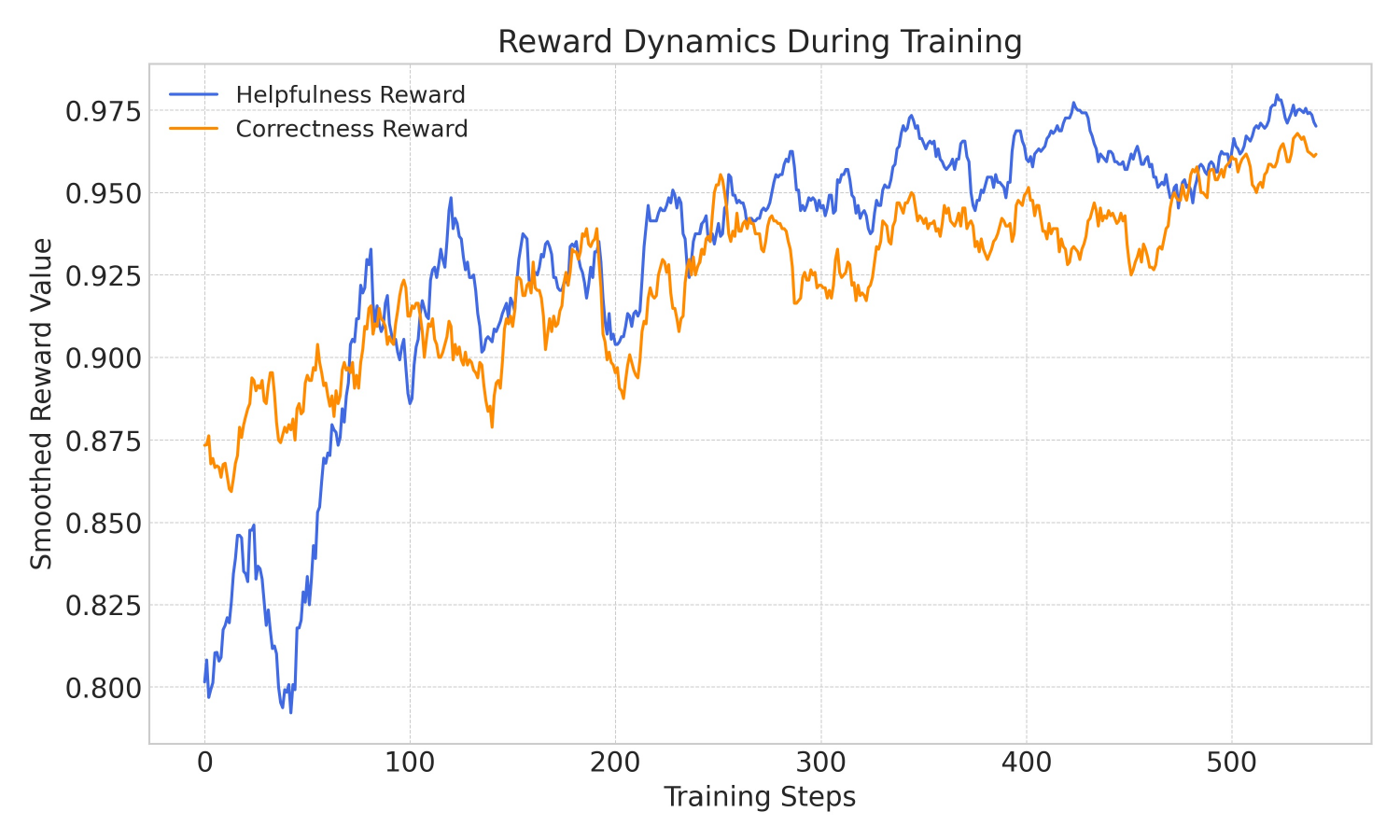}
    \caption{Training dynamics of reward components (Correctness and Helpfulness).}
    \label{fig:appendix_rewards_dynamics}
\end{figure}

\begin{figure}[htbp]
    \centering
    \includegraphics[width=0.7\linewidth]{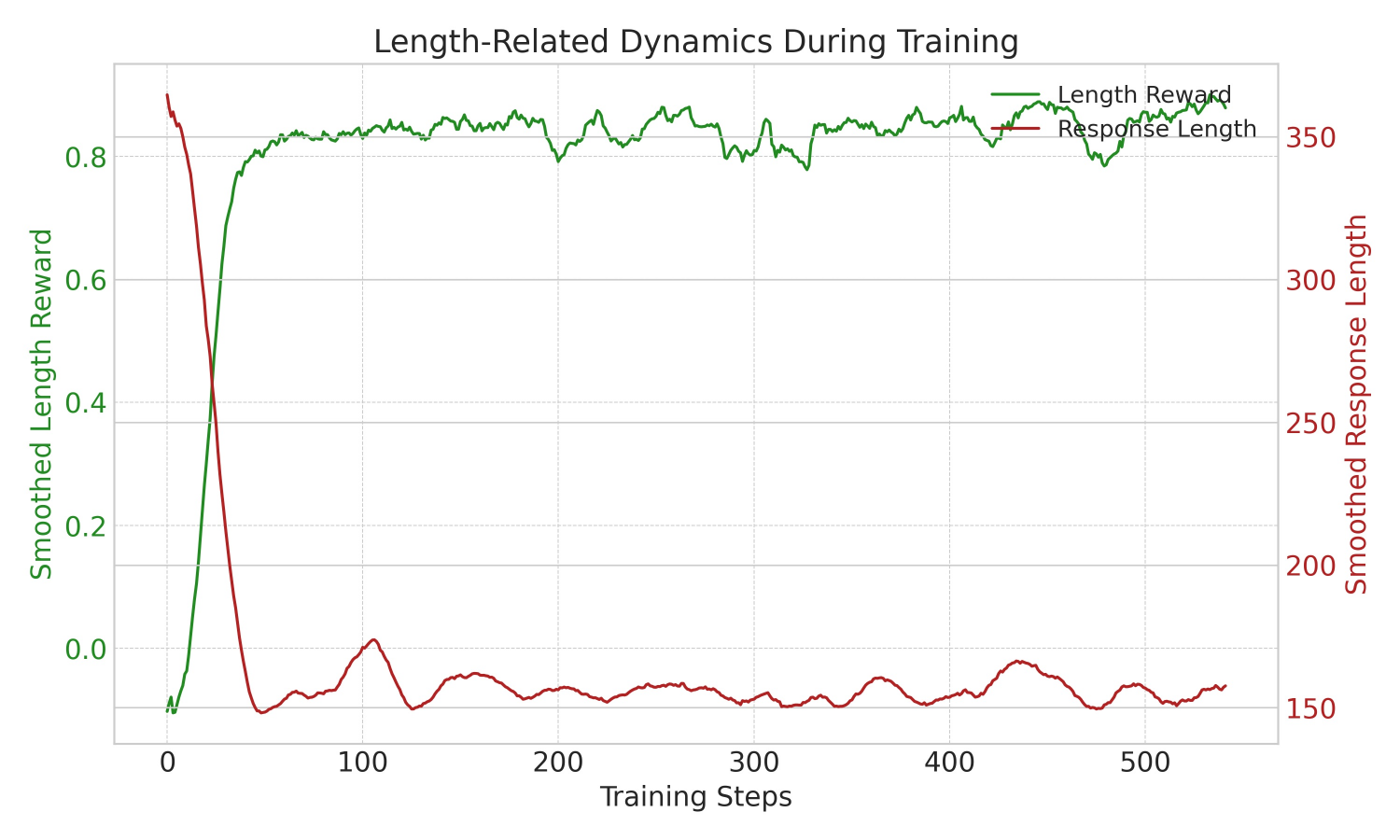}
    \caption{Training dynamics of reward components (Response Length and Length Reward).}
    \label{fig:appendix_length_rewards}
\end{figure}

\section{Prompt Templates for LLM Judge}
\label{sec:appendix_prompts}

This appendix details the prompt templates used for the LLM Judge evaluation. Table~\ref{tab:correctness_prompt_simple} outlines the template for assessing Correctness, and Table~\ref{tab:helpfulness_prompt_simple} outlines the template for assessing Helpfulness.

\begin{table*}[htbp]
  \centering
  \caption{Simplified Prompt Template for Correctness Evaluation.}
  \label{tab:correctness_prompt_simple}
  \begin{tabular}{|p{0.45\textwidth}|p{0.45\textwidth}|}
    \hline
    \textbf{Original Prompt (Chinese)} & \textbf{English Translation} \\
    \hline
    \multicolumn{2}{|c|}{\textit{Correctness LLM Judge Prompt Template}} \\
    \hline
    
    \textbf{\#\# 指令} \par
    AI主播会基于历史对话、商品信息和观众的当前问题进行回复，你需要对AI主播对观众的回复正确性进行判断。
    & 
    \textbf{\#\# Instruction} \par
    The AI streamer will respond based on historical dialogue, product information, and the  question. You need to judge the correctness of the AI streamer's response to the audience.
    \\ \hline

    \textbf{\#\# 要求} \par
    1. 判断是否正确理解观众意图，针对观众问题直接给出了回复，若答非所问应判为不正确。AI主播的复述观众部分答非所问的话也应判为不正确。回复内容能给到观众帮助的情况下，即使没有直接回复也不应算作答非所问。\par
    2. 如果有对话历史，需要考虑在对话历史下当前问题的真实意图。不能在多轮对话中答非所问，如果AI主播回复错误理解了多轮对话，那么应判为不正确。\par
    3. 回复内容既不能超出给出的商品信息，也不能出现商品信息能够回答观众问题但是AI主播不回答的情况。如果商品信息不足以回答观众问题或者观众不是问的商品信息给出的商品，那么AI主播引导观众咨询客服是正确的。\par
    4. 回复内容本身不能有逻辑矛盾的情况。\par
    5. 如果给出的信息能回答观众的部分问题，即使不能全部解决问题，如果AI主播不给出任何帮助直接引导客服，那么应该判为不正确。本条在观众重复问相同问题时不适用。\par
    6. 如果观众是在咨询关于商品属性、价格相关问题，但没有这个商品，AI主播推荐其他商品给观众是不正确的。\par
    7. 如果观众是要求主播推荐商品或者在查询商品，但是没有这个商品的这种情况下，AI主播给观众推荐其他的商品是正确的，这里的其他商品只要稍微相关就行，不需要完全符合观众要求。\par
    8. AI回复中商品属性和商品价格不能违背商品信息的属性和价格，如果与历史对话有冲突以商品信息为准。\par
    9. 问到与之前相同的问题，但是AI主播却没有提醒观众已经说过了应算作不正确，提醒观众之前已经回答过了算作正确。\par
    10. 如果观众问题和之前问题不同却提醒观众之前已经回答过了但并未针对新问题进行回复应判为不正确。\par
    11. 国补去哪领、怎么领这种问题不需要按照商品信息回答默认判为正确，但是商品本身是否有补贴需要按照商品信息进行回答。一般政府补贴是电子产品，其他产品不适用本条。\par
    12. 回复中出现``xx"这种省略词的或者最后出现话没说完的情况这种明显不通顺的都是不正确的。
    & 
    \textbf{\#\# Requirements} \par
    1. Judge whether the audience's intent was correctly understood with a direct answer provided. If the response is irrelevant, it should be judged as incorrect. The AI's paraphrasing of the question being irrelevant also counts as incorrect. However, if the response is helpful, it should not be considered irrelevant even if not a direct answer.\par
    2. If dialogue history is not empty, consider the true intent of the question within the context. If the AI's response misinterprets the multi-turn context, it should be judged as incorrect.\par
    3. The response must not exceed the provided product information, nor should it fail to answer a question when the information is available. If information is insufficient or the product is not listed, guiding the user to customer service is correct.\par
    4. The response must not contain logical contradictions.\par
    5. If the information can partially answer the question, the AI should not be judged as incorrect if it offers this partial help. This does not apply if the audience repeatedly asks the same question.\par
    6. If the audience asks about attributes/price for an unavailable product, it is incorrect for the AI to recommend other products.\par
    7. If the audience asks for a recommendation or searches for an unavailable product, it is correct for the AI to recommend other, somewhat related products.\par
    8. The attributes and prices in the response must not contradict the product information. Product information is the authoritative source in case of conflicts.\par
    9. If asked the same question again, the AI is incorrect if it doesn't mention that the question has already been answered.\par
    10. If asked a new question, the AI is incorrect if it claims the question was already answered without addressing the new question.\par
    11. Questions about claiming national subsidies are correct by default. However, a product's subsidy eligibility must be based on product information. This rule generally applies to electronics.\par
    12. Responses with placeholders like ``xx" or ending abruptly are incorrect.
    \\ \hline
    
    \textbf{\#\# 针对6、7两条的额外说明} \par \{``商品属性、价格咨询":[...], ``商品查询、推荐":[...]\}
    & 
    \textbf{\#\# Additional Clarifications for Rules 6 and 7} \par \{``Product Attribute/Price Inquiry":[...], ``Product Search/Recommendation":[...]\}
    \\ \hline
    
    \textbf{\#\# 输出示例} \par
    \{``判断原因": ``xxx", ``回复是否正确": ``xxx" \}
    & 
    \textbf{\#\# Output Example} \par
    \{``Reason for Judgment": ``xxx", ``Is the Response Correct": ``xxx" \}
    \\ \hline
    
    \textbf{--真实数据--} \par
    \{``商品信息":\{\{item\_info\}\}, ``对话历史":\{\{history\}\}, ``观众问题":\{\{query\}\} \}
    &
    \textbf{--Real Data--} \par
    \{``Product Information":\{\{item\_info\}\}, ``Dialogue History":\{\{history\}\}, ``Audience Question":\{\{query\}\} \}
    \\ \hline
  \end{tabular}
\end{table*}

\begin{table*}[h!]
  \centering
  \caption{Simplified Prompt Template for Helpfulness Evaluation.}
  \label{tab:helpfulness_prompt_simple}
  \begin{tabular}{|p{0.45\textwidth}|p{0.45\textwidth}|}
    \hline
    \textbf{Original Prompt (Chinese)} & \textbf{English Translation} \\
    \hline
    \multicolumn{2}{|c|}{\textit{Helpfulness LLM Judge Prompt Template}} \\
    \hline
    
    \textbf{\#\# 指令} \par
    AI主播会对观众的弹幕进行回复，会给你AI主播的回复，还会给你观众的弹幕以及对话历史，你需要判断回复是否给观众带来了帮助。
    & 
    \textbf{\#\# Instruction} \par
    The AI streamer will respond to the audience's comments. You will be given the AI's response, the audience's comment, and the dialogue history. You need to judge whether the response was helpful to the audience.
    \\ \hline

    \textbf{\#\# 要求} \par
    1. 要考量的是数字人主播的回复是否给观众带来的帮助，帮助性不需要考虑回复的信息是否正确。\par
    2. 在比较帮助性时需要考虑历史对话，即回复是否对观众在历史对话中的真实意图给出了帮助性。\par
    3. 没有给出任何帮助直接让观众找客服属于没有帮助的范畴。给到观众部分帮助并让观众去找客服属于有帮助。\par
    4. 答非所问也属于没有帮助。\par
    5. AI主播没有找到观众想要的商品给观众推荐了类似的商品属于有帮助，如果没找到并直接引导观众去找客服属于没有给出任何帮助。\par
    6. 是否给出帮助的关键判断依据是AI是否给到了观众不来问就不知道的信息。没找到商品、不知道xx信息属于是没有帮助的信息。
    & 
    \textbf{\#\# Requirements} \par
    1. The main consideration is whether the AI streamer's response was helpful to the audience. Helpfulness evaluation should not consider the correctness of the information.\par
    2. When assessing helpfulness, you must consider the dialogue history to determine if the response addressed the audience's true intent.\par
    3. Directly referring the audience to customer service without providing any information is considered not helpful. Offering partial information and then suggesting customer service is considered helpful.\par
    4. An irrelevant response is considered not helpful.\par
    5. If the AI cannot find the exact product the audience wants but recommends a similar one, it is considered helpful. If it fails to find the product and simply refers to customer service, it is not helpful.\par
    6. The key criterion for helpfulness is whether the AI provided information that the audience would not have known otherwise. Stating that a product could not be found or that certain information is unknown is not considered helpful.
    \\ \hline
    
    \textbf{\#\# 输出示例} \par
    \{``判断原因": ``xxx", ``回复有无帮助": ``xxx" \}
    & 
    \textbf{\#\# Output Example} \par
    \{``Reason for Judgment": ``xxx", ``Is the Response Helpful": ``xxx" \}
    \\ \hline
    
    \textbf{--真实数据--} \par
    \{``对话历史":\{\{history\}\}, ``观众问题":\{\{query\}\}, ... \}
    &
    \textbf{--Real Data--} \par
    \{``Dialogue History":\{\{history\}\}, ``Audience Question":\{\{query\}\}, ... \}
    \\ \hline
  \end{tabular}
\end{table*}

\section{Detailed Case Studies}
\label{app:case_study}

\begin{table*}[htbp]
  \caption{Case Study: Comparison of RFT and LiveThinking models for handling queries about unavailable products.}
  \label{tab:case-study-ac}
  \renewcommand{\arraystretch}{1.3} 
  \begin{tabular}{p{0.10\linewidth} p{0.75\linewidth}}
    \toprule
    \textbf{Component} & \textbf{Details \& Comparison} \\
    \midrule
    
    \textbf{Question} & 
    \textbf{Original (CN):} 推荐一款单冷的空调。\textbf{Translation (EN):} Recommend a cool-only air conditioner. \\
    
    \midrule
    
    \textbf{Simplified Product Infomation} & 
    The provided data contains three air conditioner products. A crucial attribute is `冷暖类型' (Cooling/Heating Type). \textbf{Link 17 (馨天翁):} `冷暖电辅' (Cool \& Warm with Electric Heat). \textbf{Link 264 (云欢):} `冷暖电辅' (Cool \& Warm with Electric Heat). \textbf{Link 261 (云欢):} `冷暖' (Cool \& Warm).
    \textbf{Key Finding:} None of the available products are of the ``single-cool" type requested by the viewer. \\

    \midrule

    \textbf{Methodology} & \textbf{Qwen3-30B-A3B + RFT} \\
    \midrule
    \textit{Reasoning Chain-of-Thought:} & 
    \textbf{Original (CN):} “好的，现在需要处理观众的问题...首先，我要查看提供的商品信息...浏览商品信息中的三个空调产品...所有商品的冷暖类型都是冷暖电辅或冷暖，没有单冷的型号。因此，商品信息中没有符合观众要求的单冷空调...根据要求，如果商品信息不足以回答...需要给出保守的高情商回复，不正面回答但提供情绪价值，并引导咨询客服...需要引用观众的问题...说明没有找到...保持口语化...”
    \newline
    \textbf{Translation (EN):} ``Okay, I need to process the audience's question... First, I must check the provided product information... Browsing the three air conditioner products... all products' cooling/heating types are `cool \& warm with electric heat' or `cool \& warm', there are no `single-cool' models. Therefore, the product information does not contain a product matching the audience's request... According to the requirements, if the information is insufficient... I need to give a conservative, high-EQ response, not answering directly but providing emotional value, and guide them to customer service... I need to quote the audience's question... state that nothing was found... maintain a colloquial tone..."
    \newline
    \textbf{Analysis:} The reasoning is verbose and meta-analytical. It describes the rules it needs to follow rather than simply executing the task.
    \\
    \cmidrule{2-2}
    \textit{Generated Response:} & 
    \textbf{Original (CN):} 有家人问推荐单冷空调哈，目前我这边暂时没看到单冷的型号呢，宝宝可以具体说一下链接号或者咨询客服帮你找找看哦。
    \newline
    \textbf{Translation (EN):} A family member is asking for a single-cool AC, huh. I'm currently not seeing any single-cool models here. Honey, could you provide a specific link number or ask customer service to help you find one?
    \newline
    \textbf{Analysis:} The response is polite but generic. It doesn't convey that the existing products were checked, which might leave the user wondering if the assistant even looked. \\

    \midrule

    \textbf{Methodology} & \textbf{LiveThinking} \\
    \midrule
    \textit{Reasoning Chain-of-Thought:} &
    \textbf{Original (CN):} “嗯，现在需要处理观众的问题...首先查看商品信息中的三个链接。链接17是冷暖型，链接264也是冷暖，链接261同样是冷暖。所有商品都是冷暖类型，没有单冷的。所以无法回答，需要回复否，并说明情况，引导提供链接号或咨询客服。”
    \newline
    \textbf{Translation (EN):} ``Okay, need to process the audience's question... First, check the three links in the product info. Link 17 is cool \& warm, link 264 is also cool \& warm, link 261 is also cool \& warm. All products are cool \& warm type, no single-cool. So, cannot answer, need to reply `no', explain the situation, and guide them to provide a link number or consult customer service."
    \newline
    \textbf{Analysis:} \textbf{The reasoning is direct, concise, and task-oriented.} It quickly verifies the data, reaches a logical conclusion, and outlines the next action without unnecessary meta-commentary.
    \\
    \cmidrule{2-2}
    \textit{Generated Response:} & 
    \textbf{Original (CN):} 有家人问推荐单冷空调哦，目前17号链接格力大1匹、264号云欢大1匹、261号云欢2匹都是冷暖型的哈，商品信息里没有单冷款呢。如果需要单冷的话可能需要定制或非标型号，宝宝可以再详细描述需求或咨询客服是否有库存哦。
    \newline
    \textbf{Translation (EN):} Someone's asking for a single-cool AC! Just so you know, our link \#17 Gree 1-HP, \#264 Yunhuan 1-HP, and \#261 Yunhuan 2-HP are all cool \& warm models. We don't seem to have a single-cool model in our product list right now. If you need a single-cool model, it might be a custom or special order. Honey, you could describe your needs in more detail or ask the customer service!
    \newline
    \textbf{Analysis:} \textbf{The response is superior.} It is transparent by explicitly mentioning the products it checked (\#17, \#264, \#261), demonstrating that it is grounded in the provided data. This builds trust and provides more useful context to the viewer. \\
    
    \bottomrule
  \end{tabular}
\end{table*}

\section{Reinforcement Learning Training Dynamics}
r multi-objective optimization effectively balances performance with conciseness. As shown in Figures \ref{fig:appendix_rewards_dynamics} and \ref{fig:appendix_length_rewards}, GRPO training sustains high rewards for correctness and helpfulness while simultaneously reducing generation length.

\end{CJK*} 

\end{document}